\documentclass{article}
\usepackage{amsmath,amssymb,amsfonts}
\usepackage{algorithmic}
\usepackage{graphicx}
\usepackage{textcomp}
\usepackage{multicol}

 \usepackage[final,nonatbib]{neurips_2024}

\usepackage[utf8]{inputenc} 
\usepackage[T1]{fontenc}    
\usepackage{hyperref}       
\usepackage{url}            
\usepackage{booktabs}       
\usepackage{nicefrac}       
\usepackage{microtype}      
\usepackage{xcolor}         
\usepackage{booktabs}
\usepackage{colortbl}
\usepackage[normalem]{ulem}
\useunder{\uline}{\ul}{}
\usepackage{adjustbox}

\usepackage[backend=biber,style=numeric]{biblatex}
\title{Going Beyond H\&E and Oncology: How Do Histopathology Foundation Models Perform for Multi-stain IHC and Immunology?}

\author{Amaya Gallagher-Syed 
  \And Elena Pontarini 
  \And Myles J. Lewis 
  \And Michael R. Barnes 
  \And Gregory Slabaugh \\
  Queen Mary University of London \\
  \texttt{{a.r.syed, e.pontarini, myles.lewis, m.r.barnes, g.slabaugh}@qmul.ac.uk} \\
}

\begin{document}

\maketitle

\begin{abstract}

This study evaluates the generalisation capabilities of state-of-the-art histopathology foundation models on out-of-distribution (OOD) multi-stain autoimmune Immunohistochemistry (IHC) datasets. We compare 13 feature extractor models, including ImageNet-pretrained networks, and histopathology foundation models trained on both public and proprietary data, on Rheumatoid Arthritis (RA) subtyping and Sjogren's Disease (SD) diagnostic tasks. Using a simple Attention-Based Multiple Instance Learning classifier, we assess the transferability of learned representations from cancer H\&E images to autoimmune IHC images. Contrary to expectations, histopathology-pretrained models did not significantly outperform ImageNet-pretrained models. Furthermore, there was evidence of both autoimmune feature misinterpretation and biased feature importance. Our findings highlight the challenges in transferring knowledge from cancer to autoimmune histopathology and emphasise the need for careful evaluation of AI models across diverse histopathological tasks. The code to run this benchmark is available at \href{https://github.com/AmayaGS/ImmunoHistoBench}{\texttt{https://github.com/AmayaGS/ImmunoHistoBench}}

\end{abstract}

\section{Introduction}

Recent advancements in digital pathology have led to the development of powerful foundation models trained on large-scale H\&E cancer datasets \cite{uni,bioptimus,gigapath,ctranspath,phikonv2,lunit, Vorontsov2024}. These models have shown remarkable performance in various cancer-related tasks \cite{ovarian1,ovarian2,phikonv2,uni,gigapath,bioptimus}. However, their ability to generalise to other histopathological domains, particularly immunohistochemistry (IHC) staining and autoimmune diseases, remains largely unexplored. This gap is particularly significant given the fundamental differences between cancer and autoimmune pathologies. Indeed, cancer and autoimmune diseases represent two ends of the immune spectrum, with distinct histopathological features reflecting their underlying pathogenesis \cite{Sakowska2022}. In cancer, the immune response is often suppressed or evaded, leading to uncontrolled growth of malignant cells \cite{Mangani2023}. Conversely, autoimmune diseases are characterised by an overactive immune response against self-tissues \cite{Pisetsky2023}. These fundamental differences manifest in contrasting histopathological patterns, aspects of which we exemplify in Figure \ref{intro}. 

\vspace{-3pt}

\subsection{H\&E and IHC provide complementary, yet contrasting information}

\begin{figure}[t]
    \centering
    \includegraphics[width=1\linewidth]{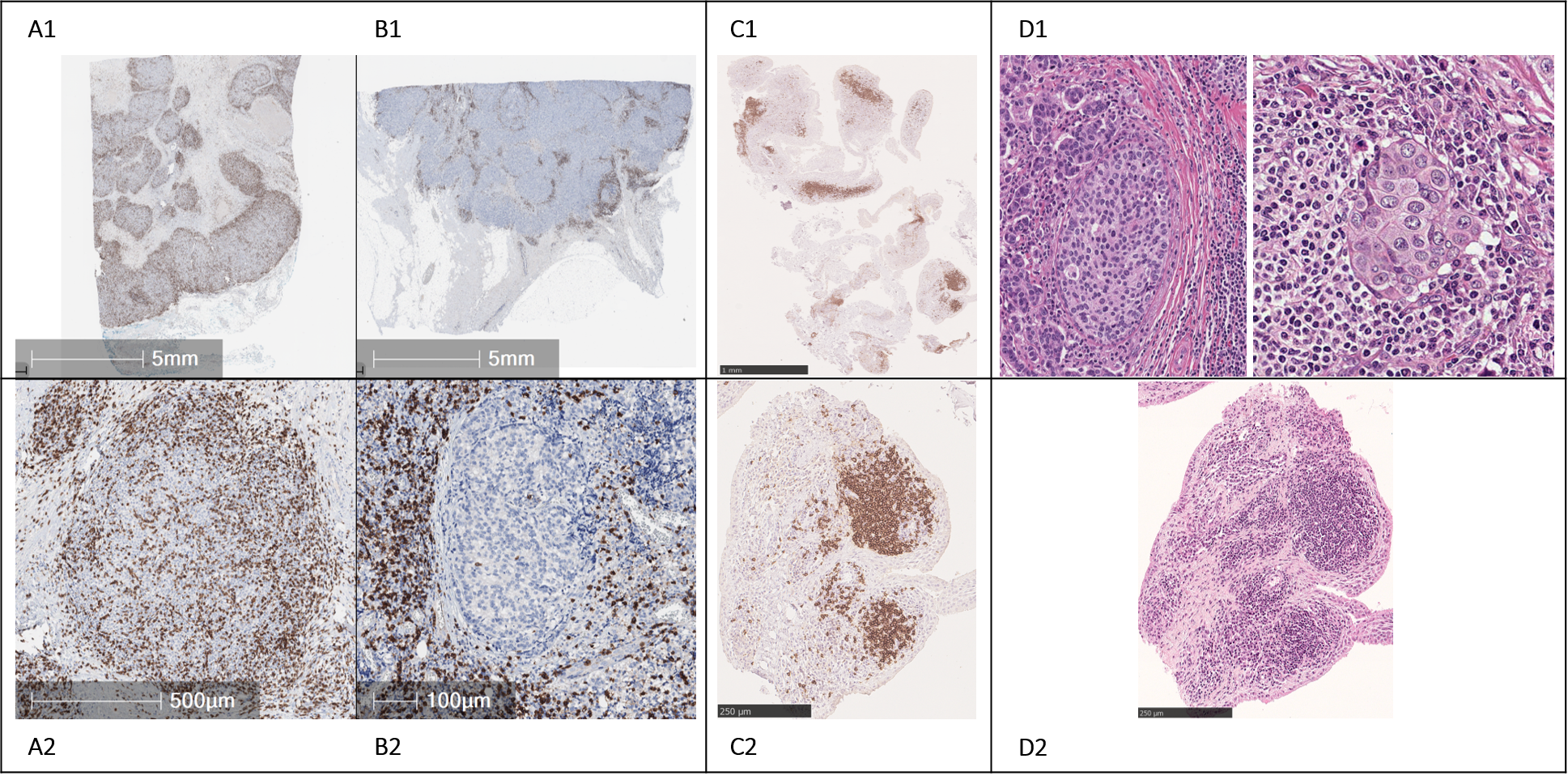}
    \caption{Differences in histological patterns between cancer and autoimmune pathologies.  \textbf{A-C}: IHC staining, \textbf{D}: H\&E staining. \textbf{A}: Breast cancer tumours (A1), heavily infiltrated with immune cells throughout (\textbf{A2}), \textbf{B}: Breast cancer tumour (\textbf{B1}), characterised by TMEs (\textbf{B2}). Tumours with immune infiltrate show better prognosis. \textbf{C}: Tissue with strong presence of immune infiltrates in autoimmune disease (\textbf{C1}) such as ELS (\textbf{C2}), which can correlate to poor prognosis. Note \textbf{A2} shows good prognosis, while \textbf{C2} shows poor. \textbf{D}: differences in H\&E staining between cancer (\textbf{D1}) and autoimmune (\textbf{D2}). In \textbf{D1}, we see examples of TMEs surrounded by more darkly stained immune cells unable to infiltrate the tumour. Cancer cells show enlarged cytoplasm to nuclear ratio. \textbf{D2} shows darkly stained immune cells.}
    \label{intro}
\end{figure}

\vspace{-3pt}

H\&E staining, a traditional and widely used technique, offers a broad view of tissue architecture and cellular morphology. Hematoxylin stains cell nuclei a deep blue-purple, while eosin stains cytoplasm and extracellular matrix in shades of pink (see Fig.\ref{intro} D). In contrast, IHC is a more specialised technique that uses antibodies tagged with visual markers to identify specific proteins or antigens within tissue samples, allowing for precise localisation and visualisation of immune cell population (see Fig.\ref{intro} A-C). In cancer diagnostics, H\&E staining remains the foundation for initial assessment and general diagnosis. However, IHC plays a crucial role in tumour classification, prognosis determination, and treatment selection by pinpointing specific cancer markers and assessing treatment responsiveness indicators. For autoimmune diseases, while H\&E staining identifies general patterns of inflammation and tissue damage, IHC becomes essential for a more nuanced understanding of the disease process. It highlights the types of immune cells present in inflammatory infiltrates, detects autoantibody deposits, and visualises specific autoantigens targeted by the immune system.

\subsection{Cancer and Autoimmunity: two sides of the immune coin}

The immune cell composition and organisation differ significantly between cancer and autoimmune conditions, reflecting their distinct pathogenic mechanisms. In H\&E-stained cancer samples, one typically observes striking alterations in cellular and tissue architecture. Cancer cells exhibit uncontrolled proliferation, leading to disorganised growth patterns and invasion into surrounding tissues. These cells often display abnormal morphology, including enlarged nuclei with irregular shapes and altered nuclear-to-cytoplasmic ratios (Fig.\ref{intro} D1). Furthermore, the immune landscape in cancers forming solid tumours is often characterized by an immunosuppressive tumour microenvironment (TME) that dampens the immune response, allowing cancer cells to evade destruction (Fig.\ref{intro} A-B). Conversely, autoimmune pathology is characterized by dense inflammatory cell infiltrates, such as Ectopic Lymphoid Structure (ELS), composed of immune cells such as lymphocytes, plasma cells, and macrophages (Fig.\ref{intro} C1/2) \cite{Mangani2023,Pitzalis2014,Sakowska2022,Elkoshi2022}. As shown in Figure \ref{intro}, these marked differences are reflected in often opposing patterns, with patterns which can indicate a good prognosis in cancer, often indicating a poor one in autoimmune pathologies.

\vspace{-3pt}

\subsection{Why does autoimmunity matter?}

These contrasting histopathological patterns reflect the underlying immunological processes at play in cancer and autoimmune diseases, providing valuable insights into their pathogenesis and potential therapeutic approaches. The intersection of autoimmunity and cancer research offers a rich landscape for knowledge discovery and therapeutic innovation. While these conditions represent opposite ends of the immune dysfunction spectrum, insights from one field often inform the other \cite{Mangani2023}. For instance, the study of immune checkpoints, originally investigated in autoimmunity, has revolutionised cancer immunotherapy \cite{Allison2015}. The presence of ELS structures, typically associated with autoimmunity, has also been linked to improved prognosis in some cancers, highlighting the complex role of immune organisation in disease outcomes \cite{Pitzalis2014,zilenaite2024intratumoral}. In turn, understanding how cancers evade immune surveillance provides clues about regulating overactive immune responses in autoimmune diseases. Importantly, knowledge gained from treating autoimmune conditions helps manage immune-related adverse events in cancer immunotherapy. This bidirectional flow of insights not only deepens our understanding of immune system dynamics but also paves the way for more nuanced and effective treatments in both cancer and autoimmune diseases. In the decades to come, novel treatment combinations and newly identified druggable targets will only expand the role of immunotherapy in the treatment of cancer and vice versa \cite{Waldman2020}.

\subsection{Challenges for H\&E generalisation}

The distinct pathogenic mechanisms, immune responses, and tissue alterations in cancer versus autoimmune diseases result in characteristically different histopathological patterns that reflect their underlying biology. Histopathology foundation models trained primarily on cancer-related H\&E data might face certain challenges when applied to autoimmune conditions and IHC analysis. We outline some potential pitfalls: 

\begin{enumerate}
    \item  \textbf{Misinterpretation of unique autoimmune features}: these models may fail to recognise or correctly interpret distinctive autoimmune patterns, such as ELSs or specific inflammatory infiltrates, leading to potential misdiagnosis.
    
    \item \textbf{Difficulty with complex IHC staining patterns}: The diverse and intricate staining patterns in IHC, especially in autoimmune diseases, could prove challenging for models trained mainly on H\&E images, resulting in inaccurate interpretation of immune cell interactions and signalling molecules.

    \item \textbf{Bias in feature importance}: These models might inappropriately recognise cancer-related features (e.g., nuclear atypia, TMEs) when analysing autoimmune conditions, potentially skewing diagnosis or disease severity assessment.
    
\end{enumerate}

\subsection{Contribution}

This study aims to assess the generalisation capabilities of state-of-the-art histopathology foundation models on out-of-distribution (OOD) multi-stain autoimmune datasets. We compare their performance against traditional ImageNet pretrained feature extraction models using simple Attention-Based Multiple Instance Learning algorithm (ABMIL) \cite{Ilse18}. This straightforward, zero-shot approach allows us to directly evaluate the transferability of learned representations from cancer H\&E images to autoimmune IHC images without the confounding effects of complex downstream tasks. It serves as an initial step to understanding if cancer-trained models can capture relevant features in autoimmune tissues, despite differences in disease processes and staining techniques.

By bridging the gap between cancer and autoimmune histopathology, this report aims to contribute to the broader goal of developing more comprehensive and versatile AI tools for medical image analysis, ultimately supporting advancements in precision medicine and improved patient care across a wider spectrum of immune-mediated diseases.

\section{Methodology}\label{sec:experiments}

\subsection{Pipeline}\label{sec:pipeline}

\begin{figure}[h]
    \centering
    \includegraphics[width=1\linewidth]{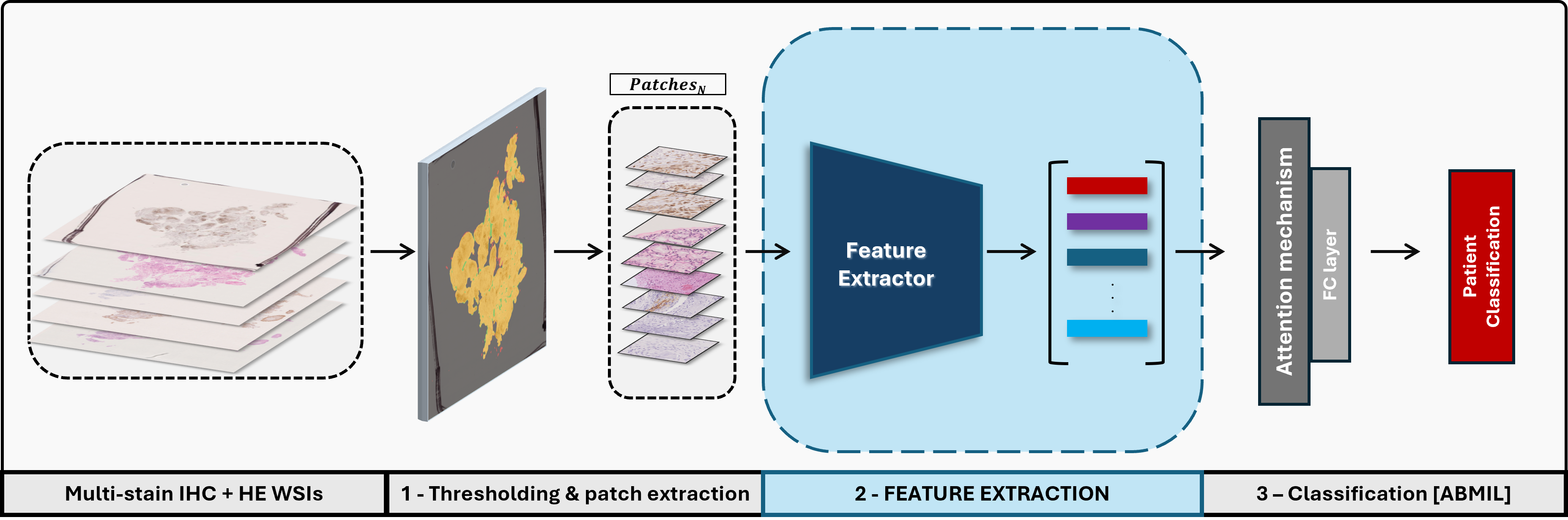}
    \caption{Schematic representation of the multi-stain whole slide image (WSI) analysis pipeline. The process encompasses (1) tissue thresholding and patch extraction, (2) feature extraction yielding patient-level feature matrices, and (3) attention-based multiple instance learning (ABMIL) classification. This study primarily focuses on comparing feature extraction methodologies.}
    \label{pipeline}
\end{figure}

Figure \ref{pipeline} illustrates our analytical pipeline. The input comprises a set of multi-stain Whole Slide Images (WSIs) per patient, including H\&E and various IHC stains listed in Table~\ref{metadata}. We apply adaptive thresholding to identify tissue areas and extract 224x224 pixel non-overlapping patches. These patches are then processed through a feature extractor network, generating a matrix of feature vectors representing each patient's WSI set. For downstream classification, we employ the Attention-Based Multiple Instance Learning (ABMIL) \cite{Ilse18} algorithm, chosen for its simplicity and strong benchmark performance. Notably, this study concentrates on comparing different feature extraction techniques within this framework.

\subsection{Datasets}

To provide a benchmark on autoimmune multi-stain datasets, we use two clinical datasets. One dataset derives from the clinical trial R4RA \cite{Humby2021}, where patients with difficult to treat Rheumatoid Arthritis (RA) were recruited for treatment with rituximab \cite{Lewis2019}. The other derives from the routine diagnostic of patients presenting with dry eyes and mouth (sicca) and investigated for Sjogren's Disease (SD). Each dataset is composed of H\&E slides, with approximately 3 IHC slides of different immune biomarkers per patient. In Table \ref{metadata}, we give further information on the stains present in each dataset. 

\begin{table}[h]
 \centering
\caption{Metadata and dataset characteristics for Sjogren's Disease and Rheumatoid Arthritis cohorts, including number of patients, WSIs, stains present and average number of stains per patient. We highlight in pink {\color[HTML]{AA336A}H\&E} staining and blue {\color[HTML]{00009B}IHC}.}
\label{metadata}
\resizebox{0.55\textwidth}{!}{%
\begin{tabular}{@{}rcccc@{}}
\toprule
\multicolumn{1}{l}{} & \multicolumn{2}{c}{\textbf{Sjogren}} & \multicolumn{2}{c}{\textbf{Rheumatoid Arthritis}} \\ \midrule
\textbf{No. Patients} & \multicolumn{2}{c}{93} & \multicolumn{2}{c}{153} \\
\textbf{No. Slides} & \multicolumn{2}{c}{347} & \multicolumn{2}{c}{607} \\
\textbf{No. Stains} & \multicolumn{2}{c}{5} & \multicolumn{2}{c}{4} \\
\textbf{Av. Stains per patient} & \multicolumn{2}{c}{3.7} & \multicolumn{2}{c}{3.97} \\
\textbf{Magnification} & \multicolumn{2}{c}{20x} & \multicolumn{2}{c}{10x} \\
\textbf{Total no. patches} & \multicolumn{2}{c}{237k} & \multicolumn{2}{c}{275k} \\
\textbf{Av. Patches per patient} & \multicolumn{2}{c}{2 530} & \multicolumn{2}{c}{1800} \\ \midrule
\textbf{Patches per stain} & \textbf{Mean} & \textbf{Total} & \textbf{Mean} & \textbf{Total} \\
{\color[HTML]{AA336A} \textbf{HE}} & 650 & 61055 & 434 & 66511 \\
{\color[HTML]{00009B} \textbf{CD3}} & 625 & 58712 & 0 & 0 \\
{\color[HTML]{00009B} \textbf{CD138}} & 377 & 35416 & 481 & 73624 \\
{\color[HTML]{00009B} \textbf{CD20}} & 626 & 58805 & 351 & 53768 \\
{\color[HTML]{00009B} \textbf{CD21}} & 254 & 23843 & 0 & 0 \\
{\color[HTML]{00009B} \textbf{CD68}} & 0 & 0 & 535 & 81915 \\ \midrule
\textbf{ML problem type} & \textbf{Detection} & \textbf{} & \textbf{Subtyping} &  \\
\textbf{Labels} & Negative & 46 & \begin{tabular}[c]{@{}c@{}}Low\\ inflammatory\end{tabular} & 66 \\
\multicolumn{1}{c}{} & Positive & 47 & \begin{tabular}[c]{@{}c@{}}High\\ inflammatory\end{tabular} & 87 \\ \bottomrule
\end{tabular}%
}
\end{table}

\subsubsection{Rheumatoid Arthritis}

\begin{figure}[h]
\centering
\vspace*{-0.1in}
\includegraphics[width=0.65\textwidth]{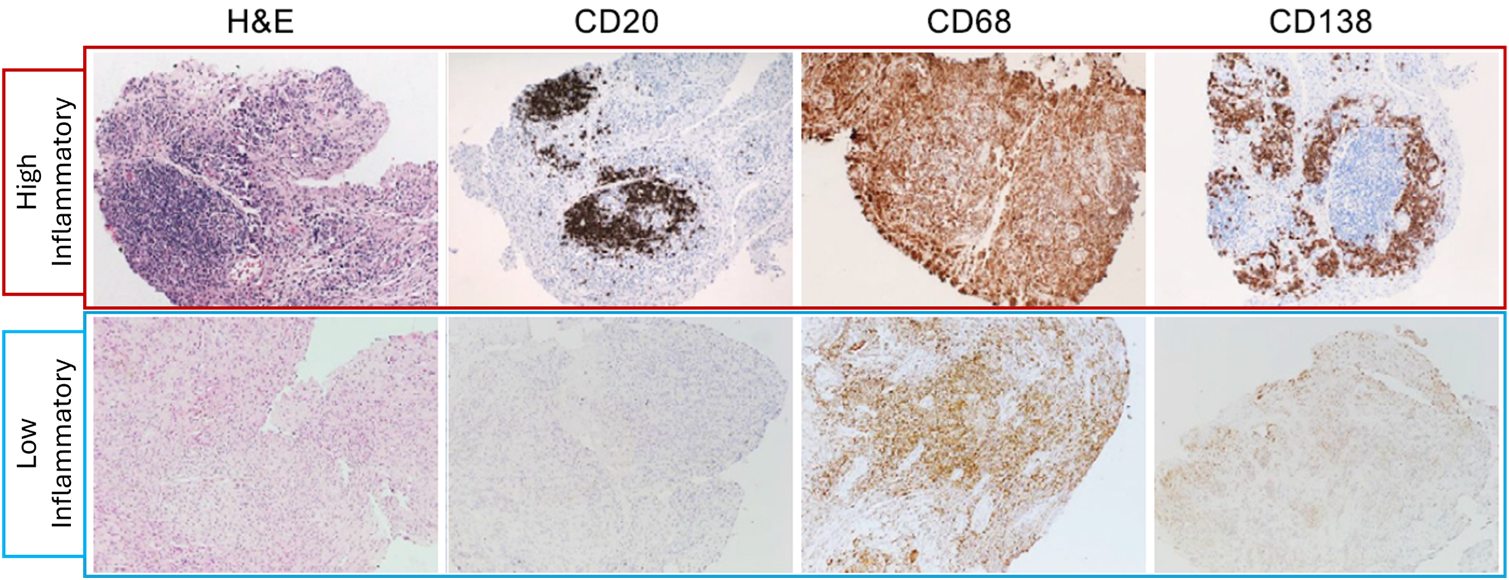}
\caption{Rheumatoid Arthritis inflammatory pathotypes based on semi-quantitative analysis of synovial tissue biopsies stained with H\&E, CD20+ B cells, CD68+ macrophages and IHC+ CD138 plasma cells \cite{Humby2021,AGS23_BMVC}.}
\label{ra_stain_types}
\end{figure}

This clinical trial dataset comprises 607 Whole Slide Images (WSIs) from 153 Rheumatoid Arthritis patients, categorised into low (N=66) and high (N=87) inflammatory subtypes \cite{Humby2021}. As exemplified in Figure \ref{ra_stain_types}, samples were stained with H\&E and the IHC markers CD20+ B cells, CD68+ macrophages and CD138+ macrophages. The dataset features a variable number of stains, averaging 3.9 per patient. We perform binary classification on low (N=66) and high (N=87) inflammatory subtypes. We extract non-overlapping patches at a 10x magnification, keeping those with over 40\% tissue coverage, totalling approximately 275k patches.

\subsubsection{Sjogren} 
\begin{figure}[h]
\centering
\vspace*{-0.1in}
\includegraphics[width=0.65\textwidth]{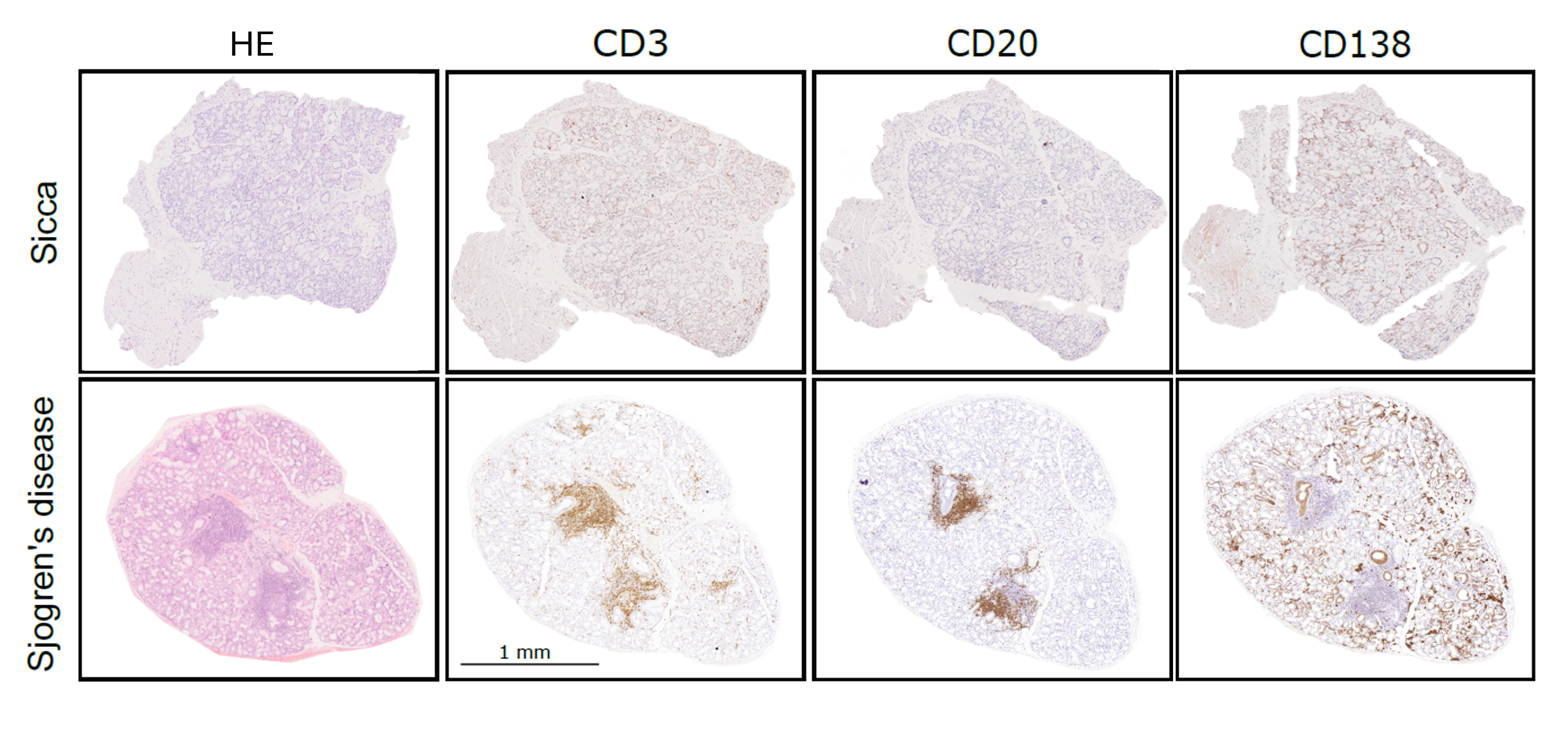}
\caption{Example of sicca vs Sjogren's Disease presentation in H\&E and IHC stains. On top, a patient diagnosed with sicca, on bottom a patient diagnosed with Sjogren's disease. Here we show samples stained with IHC stains CD3+ T cells, CD20+ B cells and CD138+ plasma cells.}
\label{sd_stain_types}
\end{figure}

Diagnostic dataset consisting of 347 WSIs of labial salivary gland biopsies sampled from 93 patients, with 46 cases of non-specific sicca and 47 cases of Sjogren's Disease (SD). As shown in Figure \ref{sd_stain_types} the samples were stained with H\&E and the IHC stains CD20+ B cells, CD3+ T cells, CD21+ follicular dendritic cell network and CD138+ plasma cells. Each patient has a variable set of multi-stain WSIs, averaging 3.7 stains per patient. We perform binary classification on the detection of sicca vs SD+ biopsies. We extract non-overlapping patches at a 20x magnification, keeping those with over 30\% tissue coverage, totalling approximately 237k patches. 

\subsection{Feature Extraction Models}

We evaluate a comprehensive and diverse set of 13 feature extractor models to assess their efficacy on Autoimmune IHC datasets. As detailed in Table \ref{methods_metadata}, these models span three distinct categories: 5 general computer vision models pretrained on ImageNet \cite{vgg16,resnet,convnext,vit}, 5 histopathology-specific foundation models trained on publicly available data SSL\_ResNet18 \cite{Ciga}, SSL\_ResNet50 \cite{lunit}, CTransPath \cite{ctranspath}, Lunit \cite{lunit} and Phikon-v2 \cite{phikonv2}, as well as three of the latest, largest models trained on proprietary data UNI, H-Optimus-0 and GigaPath \cite{uni,bioptimus,gigapath}. This selection enables us to systematically compare the performance of generic visual features against domain-specific representations learned from histopathological data. Our analysis encompasses both Convolutional Neural Network (CNN) architectures and Self-Attention (SA) based models, including state-of-the-art Vision Transformers. The models vary significantly in scale, ranging from 11 million to 1.1 billion parameters, and in their pretraining data, from generic image datasets to specialised collections of up to 1.3 billion histopathology image patches.

\begin{table}[h]
\centering
\label{methods_metadata}
\begin{adjustbox}{width=1.3\textwidth,center}
\begin{tabular}{@{}rccccccccc@{}}
\toprule
\multicolumn{1}{c}{\cellcolor[HTML]{FFFFFF}\textbf{Model}} & \textbf{\begin{tabular}[c]{@{}c@{}}Operation \\ type\end{tabular}} & \textbf{\begin{tabular}[c]{@{}c@{}}Backbone\\ Architecture\end{tabular}} & \textbf{Parameters} & \textbf{\begin{tabular}[c]{@{}c@{}}Training\\ type\end{tabular}} & \textbf{\begin{tabular}[c]{@{}c@{}}Training \\ Data Origin\end{tabular}} & \textbf{\begin{tabular}[c]{@{}c@{}}Tissue\\ types\end{tabular}} & \textbf{Magnifications} & \textbf{\begin{tabular}[c]{@{}c@{}}WSIs \\ (patches)\end{tabular}} & \textbf{Stains} \\ \midrule
\cellcolor[HTML]{FFFFFF}{\color[HTML]{E69F00} \textbf{VGG16} \cite{vgg16}} & CNN & - & 138M & Supervised & ImageNet-1k & - & - & - & - \\
\cellcolor[HTML]{FFFFFF}{\color[HTML]{E69F00} \textbf{ResNet18} \cite{resnet}} & CNN & - & 11M & Supervised & ImageNet-1k & - & - & - & - \\
\cellcolor[HTML]{FFFFFF}{\color[HTML]{E69F00} \textbf{ResNet50} \cite{resnet}} & CNN & - & 23M & Supervised & ImageNet-1k & - & - & - & - \\
\cellcolor[HTML]{FFFFFF}{\color[HTML]{E69F00} \textbf{ConvNeXt} \cite{convnext}} & CNN & - & 88M & Supervised & ImageNet-1k & - & - & - & - \\
\cellcolor[HTML]{FFFFFF}{\color[HTML]{E69F00} \textbf{ViT} \cite{vit}} & SA & - & 86M & Supervised & ImageNet-21k/1k & - & - & - & - \\
\cellcolor[HTML]{FFFFFF}{\color[HTML]{56B4E9} \textbf{SSL\_ResNet18} \cite{Ciga}} & CNN & ResNet18 & 11.1M & SimCLR & \begin{tabular}[c]{@{}c@{}}TCGA, CPTAC,\\ Multiple Public\end{tabular} & \begin{tabular}[c]{@{}c@{}}Cancer\\ Normal\end{tabular} & Multiple & \begin{tabular}[c]{@{}c@{}}25 000\\ (400K)\end{tabular} & \begin{tabular}[c]{@{}c@{}}H\&E\\ IHC\end{tabular} \\
\cellcolor[HTML]{FFFFFF}{\color[HTML]{56B4E9} \textbf{SSL\_ResNet50} \cite{lunit}} & CNN & ResNet50 & 23.5M & Barlow Twins & TCGA/Internal & Cancer & 20/40x & \begin{tabular}[c]{@{}c@{}}36 666\\ (32M)\end{tabular} & H\&E \\
\cellcolor[HTML]{FFFFFF}{\color[HTML]{56B4E9} \textbf{CTransPath} \cite{ctranspath}} & CNN-SA & SwinT & 27.5M & MoCO-v3 & TCGA/PAIP & \begin{tabular}[c]{@{}c@{}}Cancer\\ Normal\end{tabular} & 20x & \begin{tabular}[c]{@{}c@{}}32 220\\ (15M)\end{tabular} & H\&E \\
\cellcolor[HTML]{FFFFFF}{\color[HTML]{56B4E9} \textbf{Lunit} \cite{lunit}} & SA & ViT-S & 21.6M & DINO & TCGA/Internal & Cancer & 20/40x & \begin{tabular}[c]{@{}c@{}}36 666\\ (32M)\end{tabular} & H\&E \\
\cellcolor[HTML]{FFFFFF}{\color[HTML]{56B4E9} \textbf{Phikon-v2} \cite{phikonv2}} & SA & ViT-L & 307M & DINOv2 & \begin{tabular}[c]{@{}c@{}}TCGA, CPTAC, GTeX,\\ Multiple Public\end{tabular} & \begin{tabular}[c]{@{}c@{}}Cancer\\ Normal\end{tabular} & 20x & \begin{tabular}[c]{@{}c@{}}58 359 \\ (456M)\end{tabular} & H\&E \\
\cellcolor[HTML]{FFFFFF}{\color[HTML]{009E73} \textbf{H-Optimus-0} \cite{bioptimus}} & SA & ViT-G & 1.1B & DINOv2 & Internal & Cancer & 20x & 500 000 & H\&E \\
\cellcolor[HTML]{FFFFFF}{\color[HTML]{009E73} \textbf{UNI} \cite{uni}} & SA & ViT-L & 307M & DINOv2 & Internal/GTeX & \begin{tabular}[c]{@{}c@{}}Cancer\\ Normal\end{tabular} & 20x & \begin{tabular}[c]{@{}c@{}}100 426\\ (100M)\end{tabular} & H\&E \\
\cellcolor[HTML]{FFFFFF}{\color[HTML]{009E73} \textbf{GigaPath} \cite{gigapath}} & SA & ViT-G & 1.1B & DINOv2 & Internal & Cancer & 20x & \begin{tabular}[c]{@{}c@{}}171 189\\ (1.3B)\end{tabular} & H\&E \\ \bottomrule
\end{tabular}
\end{adjustbox}
\medskip
\caption{Comparative analysis of feature extractor models for digital pathology, contrasting ImageNet-pretrained models with histopathology-specific foundation models. Models are categorised into three groups: {\color[HTML]{E69F00}(orange)} general computer vision models pretrained solely on ImageNet; {\color[HTML]{56B4E9}(blue)} histopathology foundation models pretained on publicly available datasets; and {\color[HTML]{009E73}(green)} large-scale histopathology foundation models trained on proprietary datasets. This classification facilitates comparison between generic visual features and domain-specific representations. The table details backbone architecture, CNN vs Self-Attention (SA), parameter number, training paradigms and the origin, tissue and staining type of training data.}
\end{table}
\vspace{-5pt}
\subsection{Implementation Details} 

\textbf{Feature Extraction.} In total we extract 26 patient level feature representations for the RA and SD datasets, obtained from each of the 13 feature extraction networks with frozen weights. We maintain the embedding dimension of each feature extraction network. See Appendix \ref{extraction} for further details. 

\textbf{Experimental Setup and Evaluation Metrics.} We separate a random label stratified 20\% hold out test set and perform 5-fold random label stratified cross validation on the remaining data (train:val:test / 60:20:20). Models were trained for a maximum 200 epochs, with a patience set to 15 such that early stopping was called if no change was observed in either the loss, accuracy or AUC score for 15 epochs. Weights were kept for the model obtaining the best accuracy score on each validation set while ensuring there was no under-fitting or over-fitting of the models. Each of the 5 trained model was applied to the hold-out test. We report the mean and standard error (SE) of the results obtained on the hold-out test set for Accuracy, Macro F1-score, Precision, Recall, AUC and Average Precision. 

\textbf{ABMIL.} For classification we use the original one-layer ABMIL implementation \cite{Ilse18}, with input embeddings projected to a hidden layer of dimensions 128 and sigmoid activation function.

\textbf{Training schedule.} All models were trained using cross-entropy loss, with the AdamW optimizer set to $\beta_1=0.9$, $\beta_2=0.98$ and $\epsilon=10^{-9}$, with a learning rate $1 e^{-5}$ and weight decay $L_2=0.01$. No learning scheduler was used. We conducted an initial search on learning rate ($1 e^{-2}$, $1 e^{-3}$, $1 e^{-4}$, $1 e^{-5}$, settling for $1 e^{-5}$ as we observed the smoothest loss on average across models and datasets. Training was conducted on an NVidia A100 GPU (40Gb) \cite{King2017}.

\section{Results}

\subsection{Rheumatoid Arthritis}

\begin{figure}[h]
\centering
    \makebox[\textwidth][c]{\includegraphics[width=1\textwidth]{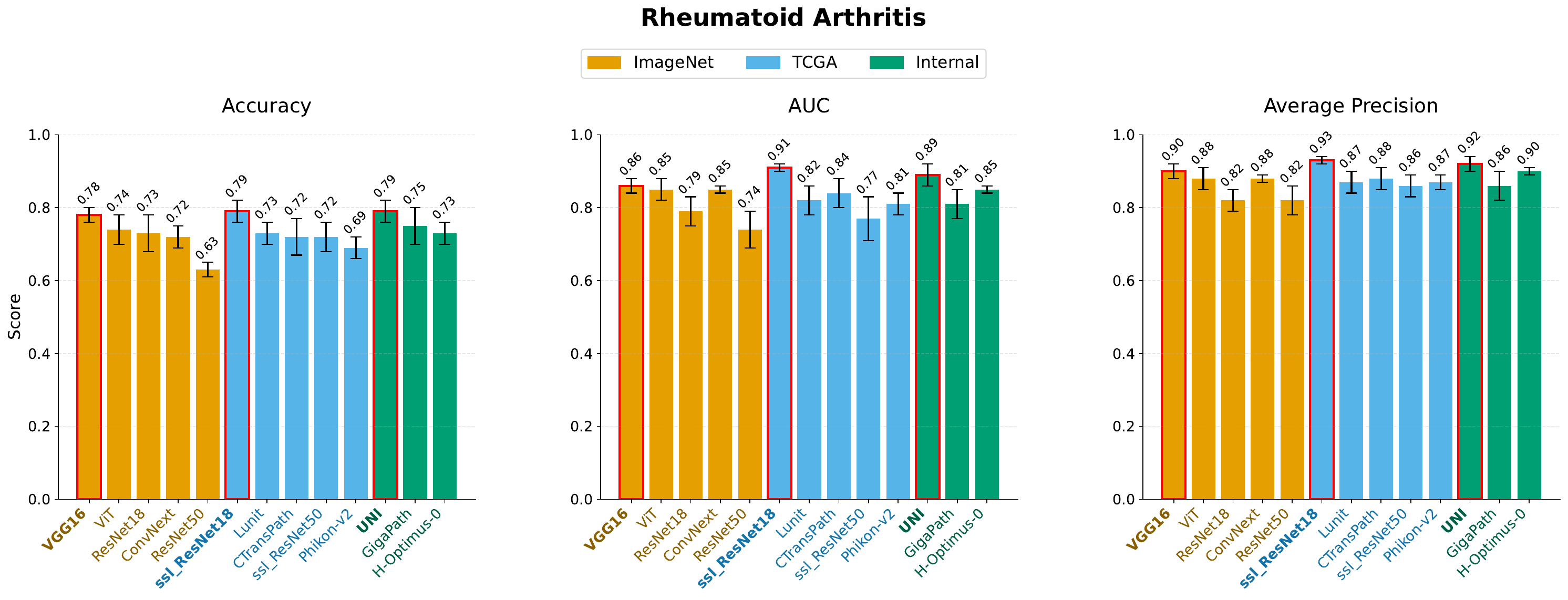}}
      \caption{Performance comparison of feature extractor models on Rheumatoid Arthritis Subtyping. ImageNet pretrained models are shown in {\color[HTML]{E69F00}orange}, Histopathology pretrained models in {\color[HTML]{56B4E9} blue} for models trained on publicly available datasets and {\color[HTML]{009E73} green} for proprietary data. Within each category and metric models are organised from most to least performant, with the top three performers highlighted with a {\fcolorbox{red}{white} {red box}}. }
  \label{ra_fig}
\end{figure}

In Figure \ref{ra_fig} and Table \ref{ra_table}, we show the results of our comparative analysis for RA, with scores organised from most to least performant within each type of model (here labelled as ImageNet, TCGA and Internal for convenience). For each metric, the top three performers are highlighted with a red box. The top three performers are SSL\_ResNet18 \cite{Ciga}, UNI \cite{uni} and VGG16 \cite{vgg16} across Accuracy, AUC and Average Precision, however there are no significant differences between the three. Overall, there are no substantial performance gain between ImageNet and histopathology-pretrained networks in general, contrary to findings from cancer-based benchmarking studies \cite{ovarian1, ovarian2, phikonv2,uni}.

\begin{table}[h]
\centering
\caption{Detailed performance metrics for various feature extraction models on Rheumatoid Arthritis classification tasks. Results include Accuracy, Macro F1-score, Precision, Recall, AUC, and Average Precision for each model, with standard errors in parentheses. Models are grouped into ImageNet-pretrained ({\color[HTML]{E69F00} orange}), publicly available histopathology-pretrained ({\color[HTML]{56B4E9} blue}), and large-scale proprietary histopathology-pretrained ({\color[HTML]{009E73} green}) categories. Top performing models highlighted in a gradient of blue.}
\label{ra_table}
\resizebox{0.9\textwidth}{!}{%
\begin{tabular}{@{}rcccccc@{}}  
\toprule
\multicolumn{1}{c}{\textbf{}} & \multicolumn{6}{c}{\textbf{Rheumatoid Arthritis}} \\ \midrule
\multicolumn{1}{l}{\textbf{}} & \textbf{Accuracy} & \textbf{\begin{tabular}[c]{@{}c@{}}Macro \\ F1-score\end{tabular}} & \textbf{Precision} & \textbf{Recall} & \textbf{AUC} & \textbf{Average Precision} \\
\textbf{\color[HTML]{E69F00} VGG16} & \cellcolor[HTML]{BBDEFE}0.78 (0.02) & \cellcolor[HTML]{BBDEFE}0.77 (0.02) & \cellcolor[HTML]{BBDEFE}0.79 (0.02) & \cellcolor[HTML]{BBDEFE}0.76 (0.01) & \cellcolor[HTML]{BBDEFE}0.86 (0.02) & \cellcolor[HTML]{BBDEFE}0.90 (0.02) \\
\textbf{\color[HTML]{E69F00} ResNet18} & 0.73 (0.05) & 0.68 (0.07) & 0.73 (0.05) & 0.69 (0.06) & 0.79 (0.04) & 0.82 (0.03) \\
\textbf{\color[HTML]{E69F00} ResNet50} & 0.63 (0.02) & 0.54 (0.05) & 0.58 (0.07) & 0.58 (0.03) & 0.74 (0.05) & 0.82 (0.04) \\
\textbf{\color[HTML]{E69F00} ConvNext} & 0.72 (0.03) & 0.65 (0.06) & 0.80 (0.02) & 0.67 (0.04) & 0.85 (0.01) & 0.88 (0.01) \\
\textbf{\color[HTML]{E69F00} ViT} & 0.74 (0.04) & 0.69 (0.06) & 0.76 (0.05) & 0.70 (0.05) & 0.85 (0.03) & 0.88 (0.03) \\
\textbf{\color[HTML]{56B4E9} ssl\_ResNet18} & \cellcolor[HTML]{84E3FF}0.79 (0.03) & \cellcolor[HTML]{84E3FF}0.76 (0.04) & \cellcolor[HTML]{84E3FF}0.86 (0.01) & \cellcolor[HTML]{84E3FF}0.76 (0.04) & \cellcolor[HTML]{84E3FF}0.91 (0.01) & \cellcolor[HTML]{84E3FF}0.93 (0.01) \\
\textbf{\color[HTML]{56B4E9} ssl\_ResNet50} & 0.72 (0.04) & 0.66 (0.07) & 0.67 (0.09) & 0.69 (0.05) & 0.77 (0.06) & 0.86 (0.03) \\
\textbf{\color[HTML]{56B4E9} CTransPath} & 0.72 (0.05) & 0.67 (0.07) & 0.77 (0.05) & 0.69 (0.05) & 0.84 (0.04) & 0.88 (0.03) \\
\textbf{\color[HTML]{56B4E9} Lunit} & 0.73 (0.03) & 0.71 (0.04) & 0.73 (0.04) & 0.71 (0.04) & 0.82 (0.04) & 0.87 (0.03) \\
\textbf{\color[HTML]{56B4E9} Phikon-v2} & 0.69 (0.03) & 0.65 (0.05) & 0.69 (0.03) & 0.67 (0.04) & 0.81 (0.03) & 0.87 (0.02) \\
\textbf{\color[HTML]{009E73} H-Optimus-0} & 0.73 (0.03) & 0.69 (0.04) & 0.77 (0.04) & 0.69 (0.03) & 0.85 (0.01) & 0.90 (0.01) \\
\textbf{\color[HTML]{009E73} UNI} & \cellcolor[HTML]{84E3FF}0.79 (0.03) & \cellcolor[HTML]{84E3FF}0.78 (0.03) & \cellcolor[HTML]{84E3FF}0.80 (0.02) & \cellcolor[HTML]{84E3FF}0.79 (0.03) & \cellcolor[HTML]{84E3FF}0.89 (0.03) & \cellcolor[HTML]{84E3FF}0.92 (0.02) \\
\textbf{\color[HTML]{009E73} GigaPath} & \cellcolor[HTML]{DAEEFE}0.75 (0.05) & \cellcolor[HTML]{DAEEFE}0.71 (0.06) & \cellcolor[HTML]{DAEEFE}0.77 (0.05) & \cellcolor[HTML]{DAEEFE}0.73 (0.06) & \cellcolor[HTML]{DAEEFE}0.81 (0.04) & \cellcolor[HTML]{DAEEFE}0.86 (0.04) \\ \bottomrule
\end{tabular}
}
\end{table}

\subsection{Sjogren's Disease}

For Sjogren's Disease, Gigapath \cite{gigapath}, ConvNext \cite{convnext} and CTransPath \cite{ctranspath} are the top three overall performers. However, as with RA we observe a similar trend of marginal differences between models, with considerable overlap in error bars and no significant differences between top performers for ImageNet vs histopathology pretrained models indicating the learned feature representation extracted from histopathology foundation models are not conferring real downstream benefits for diagnosis or subtyping of autoimmune diseases.

\begin{figure}
    \centering
    \includegraphics[width=1\linewidth]{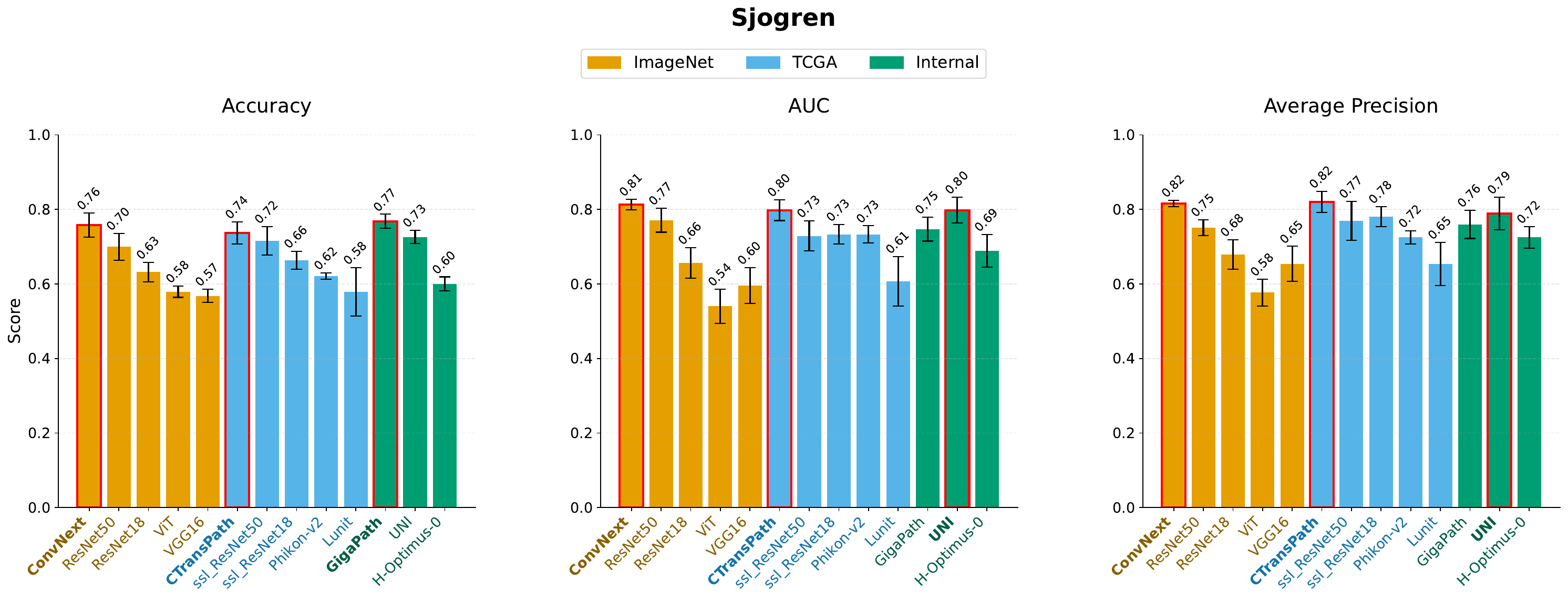}
    \caption{Performance comparison of feature extractor models on Rheumatoid Arthritis Subtyping and Sjogren's Disease Detection tasks. ImageNet pretrained models are shown in {\color[HTML]{E69F00}orange}, Histopathology pretrained models in {\color[HTML]{56B4E9} blue} for models trained on publicly available datasets and {\color[HTML]{009E73} green} for proprietary data.Within each category and metric models are organised from most to least performant, with the top three performers highlighted with a {\fcolorbox{red}{white} {red box}}. }
    \label{sjogren_fig}
\end{figure}

\begin{table}[h]
\centering
\caption{Detailed performance metrics for various feature extraction models on Sjogren's Disease classification tasks. Results include Accuracy, Macro F1-score, Precision, Recall, AUC, and Average Precision for each model, with standard errors in parentheses. Models are grouped into ImageNet-pretrained ({\color[HTML]{E69F00} orange}), publicly available histopathology-pretrained ({\color[HTML]{56B4E9} blue}), and large-scale proprietary histopathology-pretrained ({\color[HTML]{009E73} green}) categories. Top performing models highlighted in a gradient of blue.}
\label{sjogren_table}
\resizebox{0.9\textwidth}{!}{%
\begin{tabular}{@{}rcccccc@{}}  
\toprule
\multicolumn{1}{c}{\textbf{}} & \multicolumn{6}{c}{\textbf{Sjogren}} \\ \midrule
\multicolumn{1}{l}{\textbf{}} & \textbf{Accuracy} & \textbf{\begin{tabular}[c]{@{}c@{}}Macro \\ F1-score\end{tabular}} & \textbf{Precision} & \textbf{Recall} & \textbf{AUC} & \textbf{Average Precision} \\
\textbf{\color[HTML]{E69F00} VGG16} & 0.57 (0.02) & 0.57 (0.02) & 0.57 (0.02) & 0.57 (0.02) & 0.60 (0.05) & 0.65 (0.05) \\
\textbf{\color[HTML]{E69F00} ResNet18} & 0.63 (0.03) & 0.62 (0.03) & 0.63 (0.03) & 0.63 (0.03) & 0.66 (0.04) & 0.68 (0.04) \\
\textbf{\color[HTML]{E69F00} ResNet50} & 0.70 (0.03) & 0.68 (0.04) & 0.71 (0.03) & 0.69 (0.04) & 0.77 (0.03) & 0.82 (0.02) \\
\textbf{\color[HTML]{E69F00} ConvNext} & \cellcolor[HTML]{BBDEFE} 0.76 (0.03) & \cellcolor[HTML]{BBDEFE}0.75 (0.04) & \cellcolor[HTML]{BBDEFE}0.78 (0.03) & \cellcolor[HTML]{BBDEFE}0.75 (0.03) & \cellcolor[HTML]{BBDEFE}0.81 (0.01) & \cellcolor[HTML]{BBDEFE}0.82 (0.01) \\
\textbf{\color[HTML]{E69F00} ViT} & 0.58 (0.01) & 0.55 (0.03) & 0.63 (0.04) & 0.58 (0.02) & 0.54 (0.05) & 0.58 (0.04) \\
\textbf{\color[HTML]{56B4E9} ssl\_ResNet18} & 0.66 (0.02) & 0.66 (0.02) & 0.67 (0.02) & 0.66 (0.02) & 0.73 (0.03) & 0.78 (0.03) \\
\textbf{\color[HTML]{56B4E9} ssl\_ResNet50} & 0.72 (0.04) & 0.71 (0.04) & 0.72 (0.04) & 0.72 (0.04) & 0.73 (0.04) & 0.77 (0.05) \\
\textbf{\color[HTML]{56B4E9} CTransPath} & \cellcolor[HTML]{DAEEFE}0.74 (0.03) & \cellcolor[HTML]{DAEEFE}0.73 (0.03) & \cellcolor[HTML]{DAEEFE}0.76 (0.03) & \cellcolor[HTML]{DAEEFE}0.74 (0.03) & \cellcolor[HTML]{DAEEFE}0.80 (0.03) & \cellcolor[HTML]{DAEEFE}0.82 (0.03) \\
\textbf{\color[HTML]{56B4E9} Lunit} & 0.58 (0.06) & 0.57 (0.07) & 0.57 (0.07) & 0.58 (0.07) & 0.61 (0.07) & 0.65 (0.06) \\
\textbf{\color[HTML]{56B4E9} Phikon-v2} & 0.62 (0.01) & 0.62 (0.01) & 0.64 (0.02) & 0.62 (0.01) & 0.73 (0.02) & 0.72 (0.02) \\
\textbf{\color[HTML]{009E73} H-Optimus-0} & 0.60 (0.02) & 0.59 (0.02) & 0.62 (0.02) & 0.61 (0.02) & 0.69 (0.04) & 0.73 (0.03) \\
\textbf{\color[HTML]{009E73} UNI} & 0.73 (0.02) & 0.72 (0.02) & 0.74 (0.02) & 0.73 (0.02) & 0.80 (0.03) & 0.79 (0.04) \\
\textbf{\color[HTML]{009E73} GigaPath} & \cellcolor[HTML]{84E3FF} 0.77 (0.02) & \cellcolor[HTML]{84E3FF}0.77 (0.02) & \cellcolor[HTML]{84E3FF}0.77 (0.02) & \cellcolor[HTML]{84E3FF}0.77 (0.02) & \cellcolor[HTML]{84E3FF}0.75 (0.03) & \cellcolor[HTML]{84E3FF}0.76 (0.04) \\ \bottomrule
\end{tabular}
}
\end{table}

\subsection{Attention heatmaps highlight misinterpretation and bias in feature importance}

\begin{figure}[t]
    \centering
    \includegraphics[width=1\linewidth]{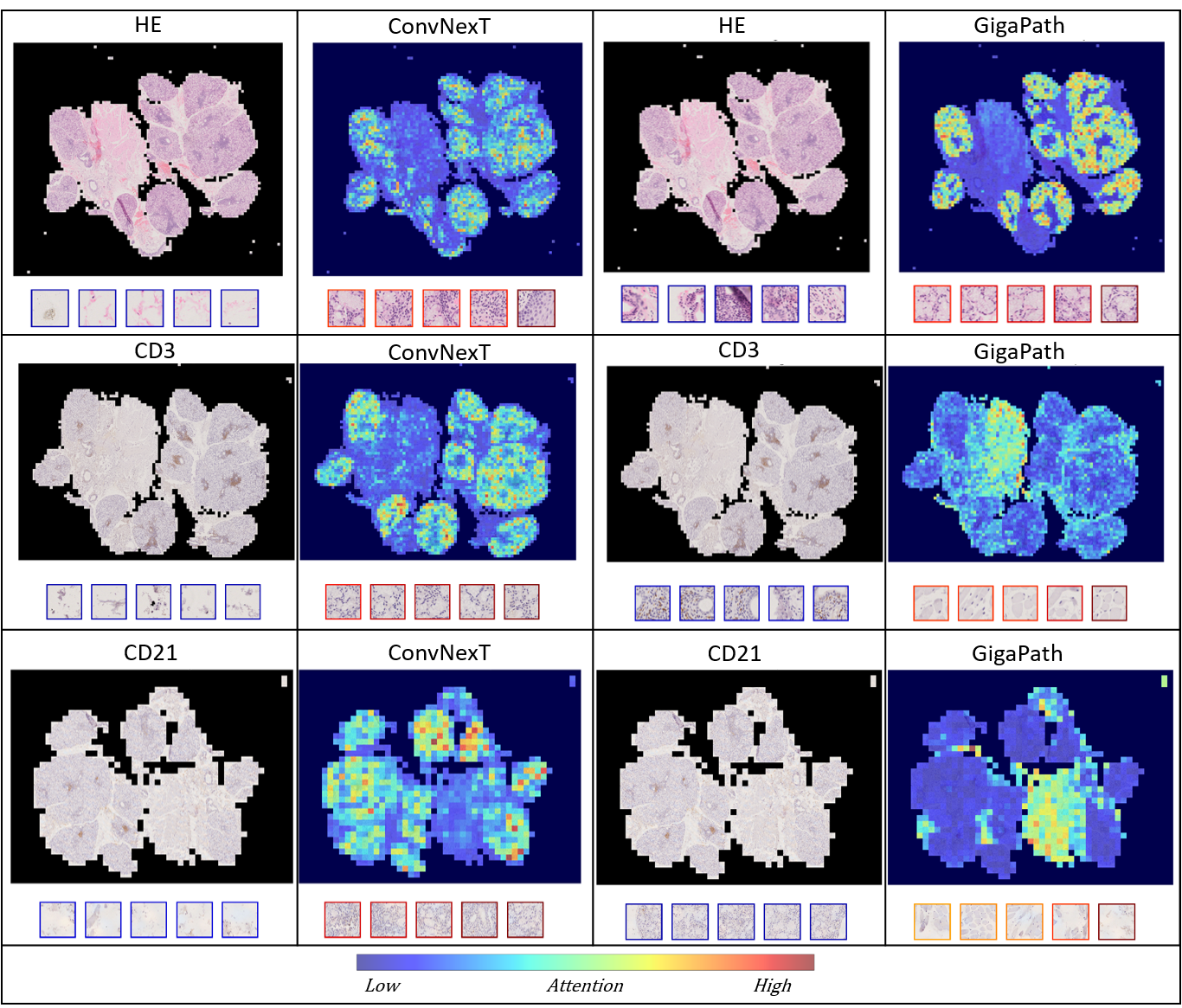}
    \caption{Attention heatmaps obtained from the ABMIL models trained with features extracted from ConvNext (left) and GigaPath (right) for a multi-stain (H\&E, CD3 and CD21) set of SD+ WSIs. These results highlight bias in feature importance, with immune cell populations being less attended to by the foundation model, as well as misinterpretation of features, with the foundation model concentrating instead on areas which bear a high level resemblance to TMEs and abnormal cancer cells. See Appendix \ref{convnext} \ref{gigapath} \ref{ctranspath} for higher resolution images.}
    \label{heatmaps}
\end{figure}

In Figure \ref{heatmaps} we show an example of a SD+ set of multi-stain WSIs (H\&E, CD3 and CD21), with attention heatmaps obtained from the ABMIL models trained with features extracted from ConvNext (left) and GigaPath (right), the two top performer models for SD. Observing these heatmaps we see clear evidence of the pitfalls mentioned in the introduction: in all three WSIs we see inflammatory aggregates present throughout the tissue being explicitly less attended to by the GigaPath foundation model. This is particularly clear in the H\&E stained slide, were for GigaPath we see inflammatory aggregates appear as darker areas of low attention scores in the middle of highly attented areas. In contrast, ConvNext has a more evenly spread out attention map covering areas of immune cell infiltration. In CD3 and CD21, we again see this pattern of more spread out attention covering areas with immune cells in ConvNexT, whereas GigaPath is picking up strongly on areas which appear to bear high level similarities to TMEs, but which are not structures present in autoimmune disease. This is particularly striking in the top 5 patches with highest and lowest attentions scores plotted below each WSIs: GigaPath consistently assigns low attention scores to areas with immune cell populations and high attention scores to tissue with morphologies which bear some resemblance to abnormal cancer cells, while ConvNexT more accurately assigns low attention scores to areas with little tissue coverage or with WSIs artefacts present and higher attention scores to areas with immune cell populations. We further show attention heatmaps for UNI, CTransPath, ResNet50, ssl\_ResNet50, Phikon-v2 and H-Optimus-0 in appendix \ref{phikon} \ref{bioptimus} \ref{ctranspath} \ref{uni} \ref{resnet50} \ref{ssl_resnet50}. Notably, UNI, CTransPath, ResNet50 and ssl\_ResNet50 show similar attention patterns, with attention scores very focused on aggregates of immune cell populations, whereas H-Optimus-0 and Phikon-v2 align with the patterns seen in GigaPath. This could indicate GigaPath, H-Optimus-0 and Phikon-v2 are focusing more on cancer specific patterns due to their large, highly specialised training corpus and training methodology. 

\paragraph{Recommendation.} For researchers working on IHC and non-cancer datasets wishing to use a histopathology foundation model, UNI and CTransPath currently seem to represent a good compromise. However, we recommend caution and careful checking of the features highlighted by different feature extractors. 

\paragraph{Limitations.} In this study we have empirically shown there is little to no significant difference in performance between ImageNet and histopathology-pretrained feature extraction networks for autoimmune IHC datasets, as well as clear differences in feature importance and heatmaps obtained by different models. However, we do not pinpoint which differences between backbone, training data and model training are the main contributors to the differences observed.

\section{Conclusion}
 
Our findings highlight the complexity of transferring knowledge from cancer histopathology to the autoimmune context and underscores the need for careful evaluation of model performance across diverse histopathological tasks, as this will be key for use in clinical practice. We believe future work should focus on incorporating a broader scope of immune-mediated diseases into histopathological foundation models, and handling complex IHC staining patterns to develop more versatile and accurate AI tools for a broader spectrum of diseases. 

By understanding how the immune system behaves in these different contexts, we can gain deeper insights into immune evasion, surveillance, and modulation. This knowledge not only bridges the gap between cancer biology and immunology but also opens new avenues for disease prevention, diagnosis, and treatment in both fields. Future AI models in digital pathology should be designed to capture and interpret these complex immunological features across different disease states, enhancing our ability to leverage insights from both autoimmunity and cancer research.

\section*{Acknowledgments and Disclosure of Funding}

We wish to thank Dr. Dovile Zilenaite for her insightful comments, knowledge and help in making Figure 1. A.G.S. receives funding from the Wellcome Trust [218584/Z/19/Z].  This paper utilised Queen Mary’s Andrena HPC facility. This work also acknowledges the support of the National Institute for Health and Care Research Barts Biomedical Research Centre (NIHR203330), a delivery partnership of Barts Health NHS Trust, Queen Mary University of London, St George’s University Hospitals NHS Foundation Trust and St George’s University of London.

\medskip

\printbibliography


\appendix

\section{Supplemental material}

\subsection{Model feature size \& extraction time}

\begin{table}[h]
\centering
\caption{We show vector feature size and extraction time, as well as peak RAM and VRAM usage for each model on both datasets.}
\label{extraction}
\resizebox{1\textwidth}{!}{%
\begin{tabular}{@{}rccccccc@{}}
\toprule
\multicolumn{1}{l}{} & \textbf{\begin{tabular}[c]{@{}c@{}}Feature vector\\ size\end{tabular}} & \multicolumn{3}{c}{\textbf{Sjogren}} & \multicolumn{3}{c}{\textbf{RA}} \\ \midrule
\multicolumn{1}{l}{} & \multicolumn{1}{l}{} & \textbf{\begin{tabular}[c]{@{}c@{}}Extraction time\\ (m)\end{tabular}} & \textbf{\begin{tabular}[c]{@{}c@{}}Peak RAM\\ (GB)\end{tabular}} & \textbf{\begin{tabular}[c]{@{}c@{}}Peak VRAM \\ (GB)\end{tabular}} & \textbf{\begin{tabular}[c]{@{}c@{}}Extraction time \\ (m)\end{tabular}} & \textbf{\begin{tabular}[c]{@{}c@{}}Peak RAM \\ (GB)\end{tabular}} & \textbf{\begin{tabular}[c]{@{}c@{}}Peak VRAM \\ (GB)\end{tabular}} \\ \midrule
\color[HTML]{E69F00} \textbf{VGG16} \cite{vgg16} & 4096 & 39 & 40 & 1.7 & 49 & 45 & 1.7 \\
\color[HTML]{E69F00} \textbf{ResNet18} \cite{resnet} & 512 & 38 & 40 & 1.2 & 43 & 33 & 1.2 \\
\color[HTML]{E69F00} \textbf{ResNet50} \cite{resnet} & 2048 & 59 & 20 & 1.2 & 64 & 45 & 1.2 \\
\color[HTML]{E69F00} \textbf{ConvNeXt} \cite{convnext} & 1024 & 99 & 22 & 2.7 & 118 & 37 & 2.6 \\
\color[HTML]{E69F00} \textbf{ViT} \cite{vit} & 768 & 79 & 45 & 0.4 & 87 & 45 & 0.4 \\
\color[HTML]{56B4E9} \textbf{SSL\_ResNet18} \cite{Ciga} & 512 & 39 & 40 & 1.2 & 43 & 34 & 1.2 \\
\color[HTML]{56B4E9} \textbf{SSL\_ResNet50} \cite{lunit} & 2048 & 57 & 38 & 1.2 & 64 & 30 & 1.2 \\
\color[HTML]{56B4E9} \textbf{CTransPath} \cite{ctranspath} & 768 & 88 & 34 & 0.3 & 99 & 40 & 0.3 \\
\color[HTML]{56B4E9} \textbf{Lunit} \cite{lunit} & 384 & 60 & 46 & 0.1 & 99 & 40 & 0.3 \\
\color[HTML]{56B4E9} \textbf{Phikon-v2} \cite{phikonv2} & 1024 & 156 & 37 & 1.2 & 190 & 50 & 1.2 \\
\color[HTML]{009E73} \textbf{H-Optimus-0} \cite{bioptimus} & 1536 & 300 & 64 & 4.6 & 336 & 50 & 4.6 \\
\color[HTML]{009E73} \textbf{UNI} \cite{uni} & 1024 & 123 & 43 & 1.2 & 150 & 50 & 1.2 \\
\color[HTML]{009E73} \textbf{GigaPath} \cite{gigapath} & 1536 & 300 & 64 & 4.6 & 325 & 52 & 4.6 \\ \bottomrule
\end{tabular}
}
\end{table}

\subsection{Attention heatmaps}

\begin{figure}
    \centering
    \includegraphics[width=1\linewidth]{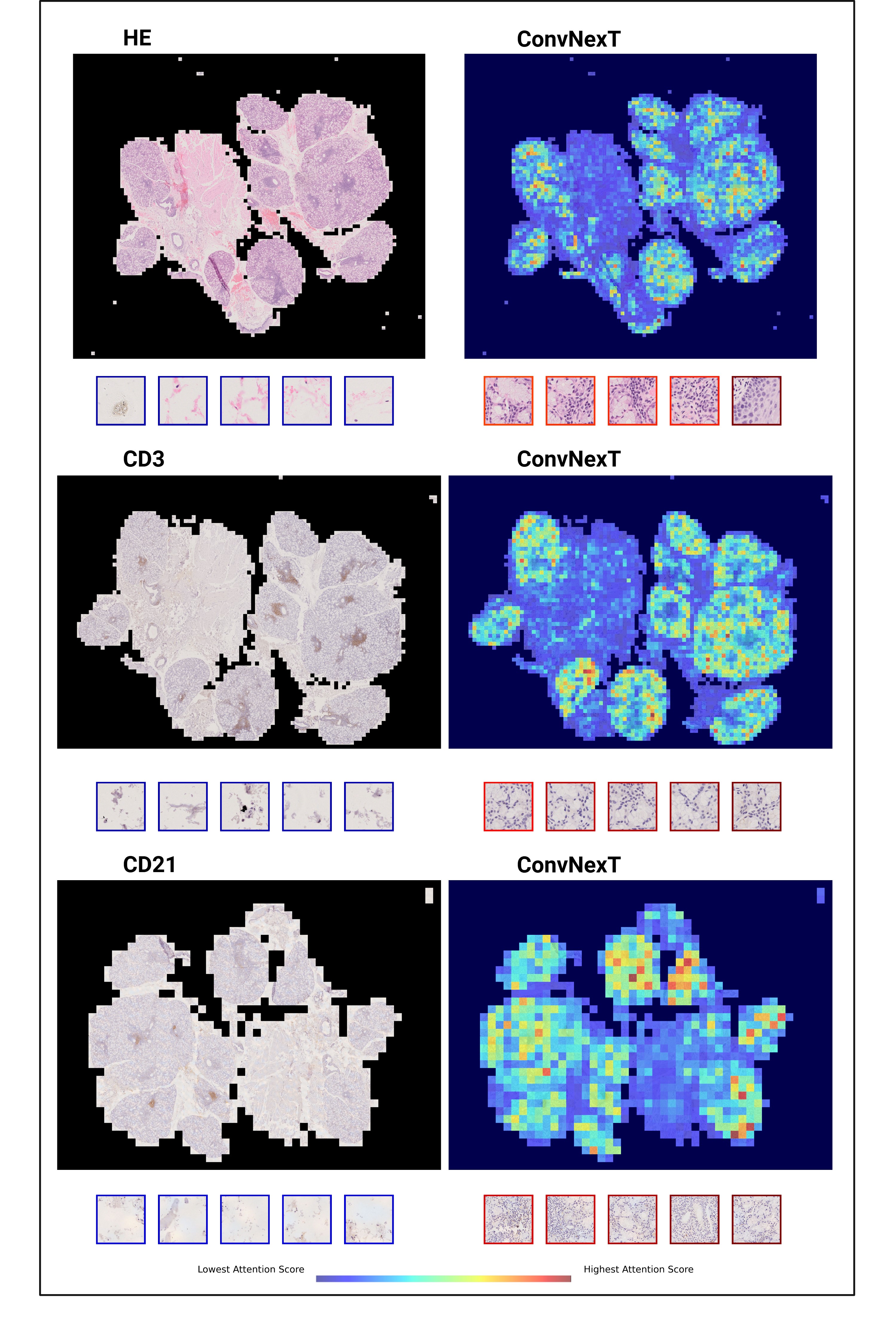}
    \caption{Attention heatmaps obtained from the ABMIL models trained with features extracted from ConvNext for a multi-stain (H\&E, CD3 and CD21) set of SD+ WSIs. Overall we see a more spread out attention pattern, covering areas of immune cell populations.}
    \label{convnext}
\end{figure}

\begin{figure}
    \centering
    \includegraphics[width=1\linewidth]{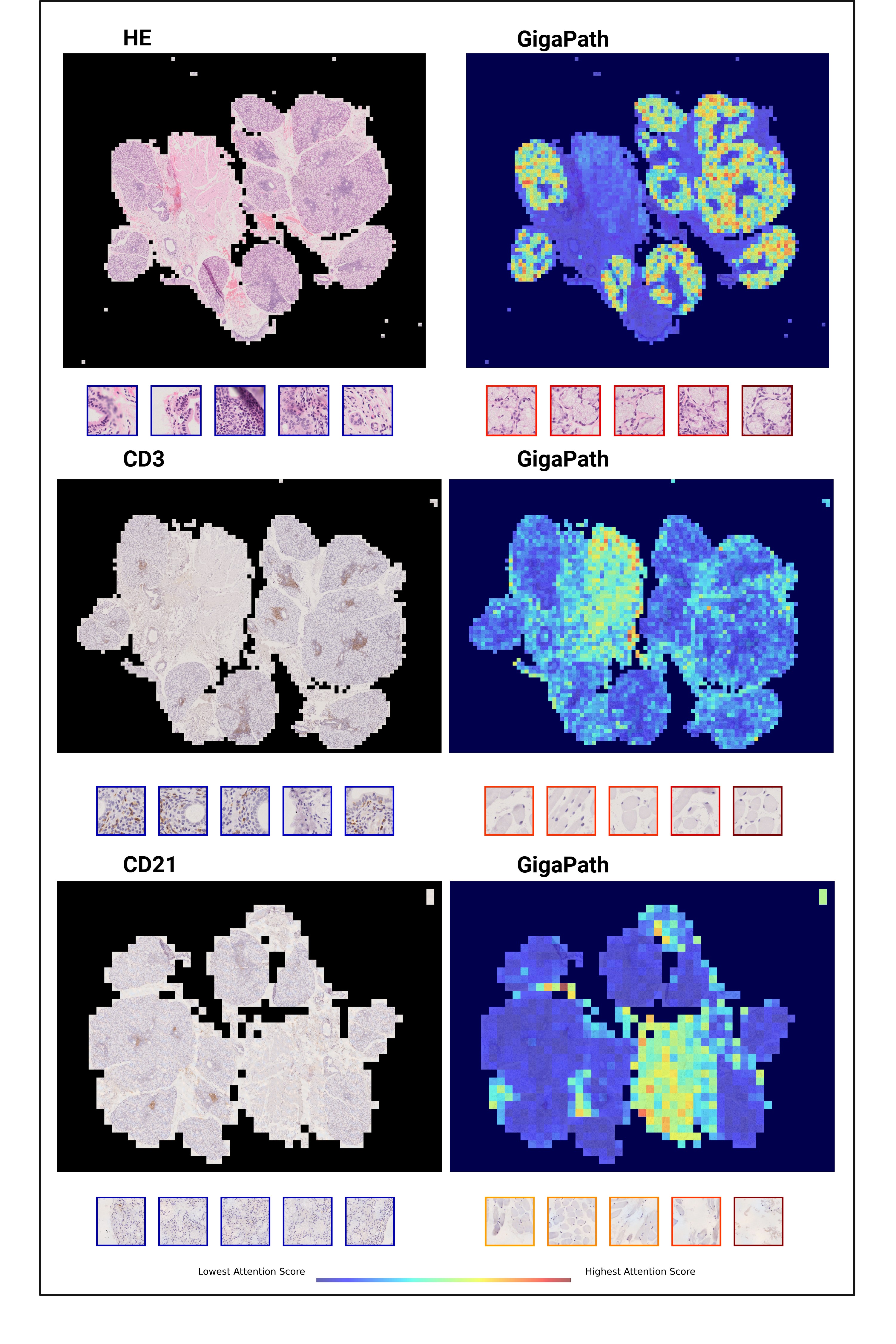}
    \caption{Attention heatmaps obtained from the ABMIL models trained with features extracted from GigaPath for a multi-stain (H\&E, CD3 and CD21) set of SD+ WSIs. We notice attention score tend to cluster more tightly around areas with no inflammatory aggregates.}
    \label{gigapath}
\end{figure}

We show higher resolution images for ConvNexT and GigaPath in \ref{convnext} and \ref{gigapath}, as well as further heatmaps for CTransPath and UNI models in \ref{ctranspath} and \ref{uni}. 

\begin{figure}
    \centering
    \includegraphics[width=1\linewidth]{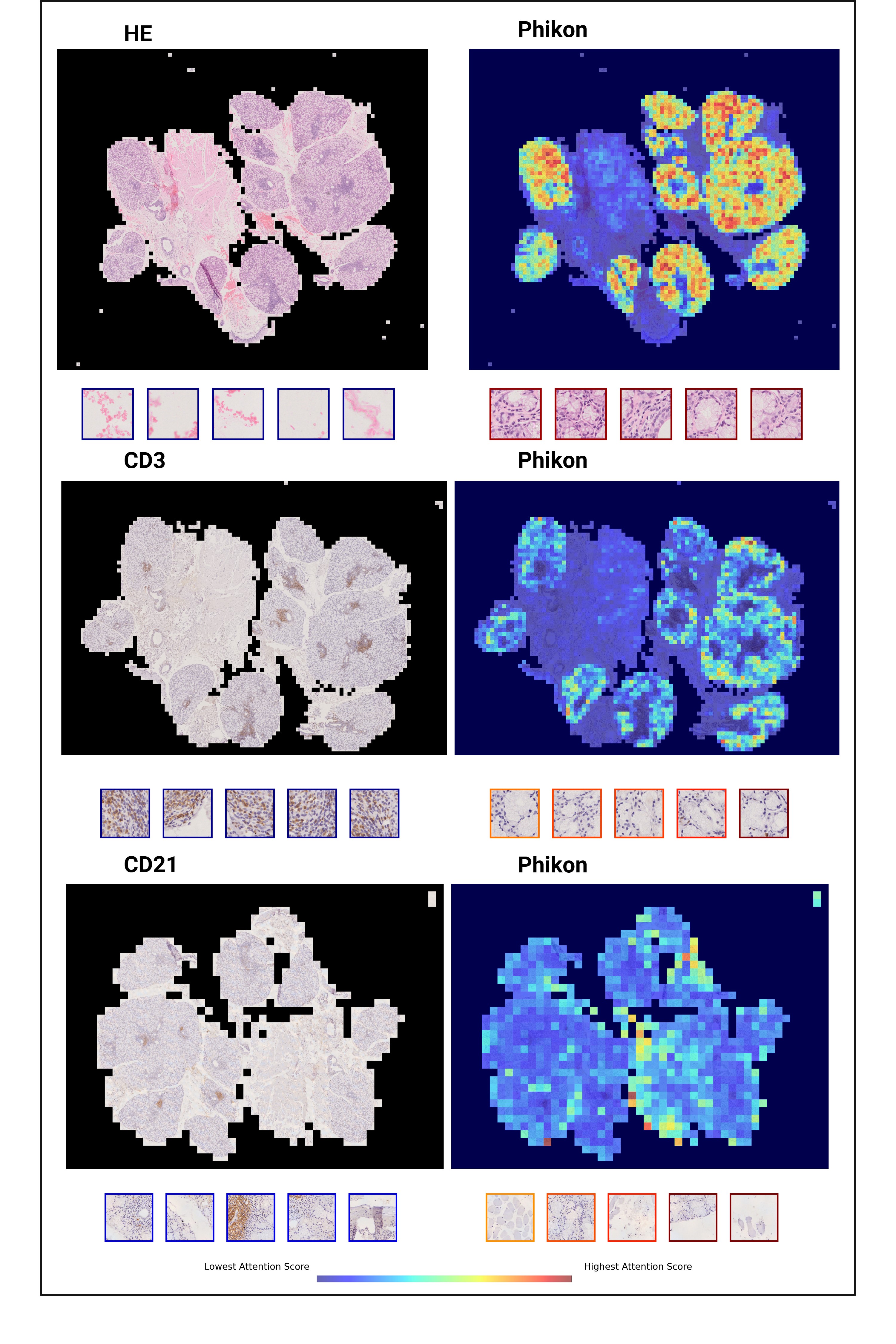}
    \caption{Attention heatmaps obtained from the ABMIL models trained with features extracted from Phikon for a multi-stain (H\&E, CD3 and CD21) set of SD+ WSIs. Attention scores patterns align with those seen in GigaPath.}
    \label{phikon}
\end{figure}

\begin{figure}
    \centering
    \includegraphics[width=1\linewidth]{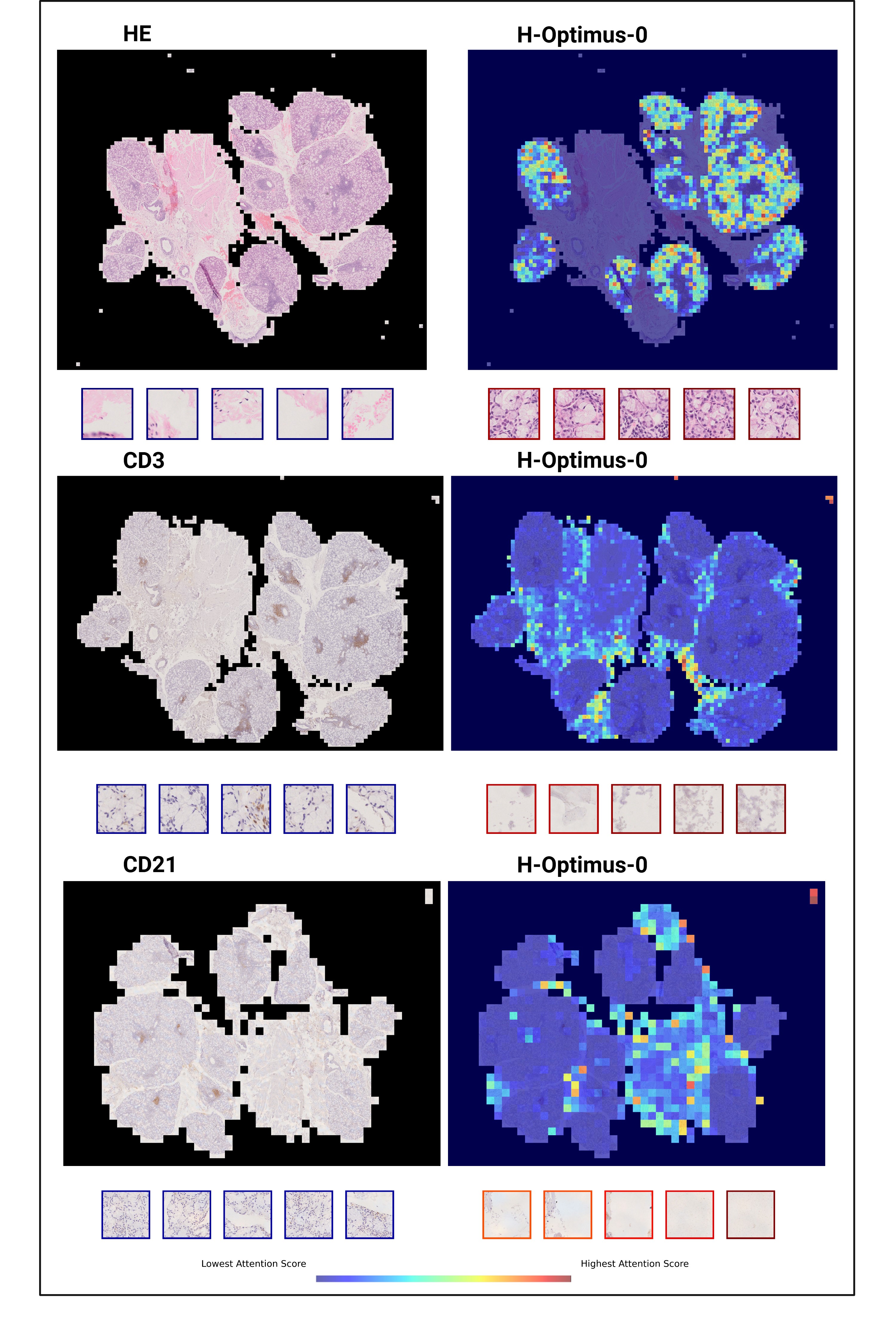}
    \caption{Attention heatmaps obtained from the ABMIL models trained with features extracted from H-Optimus-0 for a multi-stain (H\&E, CD3 and CD21) set of SD+ WSIs. Attention scores patterns align with those seen in GigaPath.}
    \label{bioptimus}
\end{figure}

\begin{figure}
    \centering
    \includegraphics[width=1\linewidth]{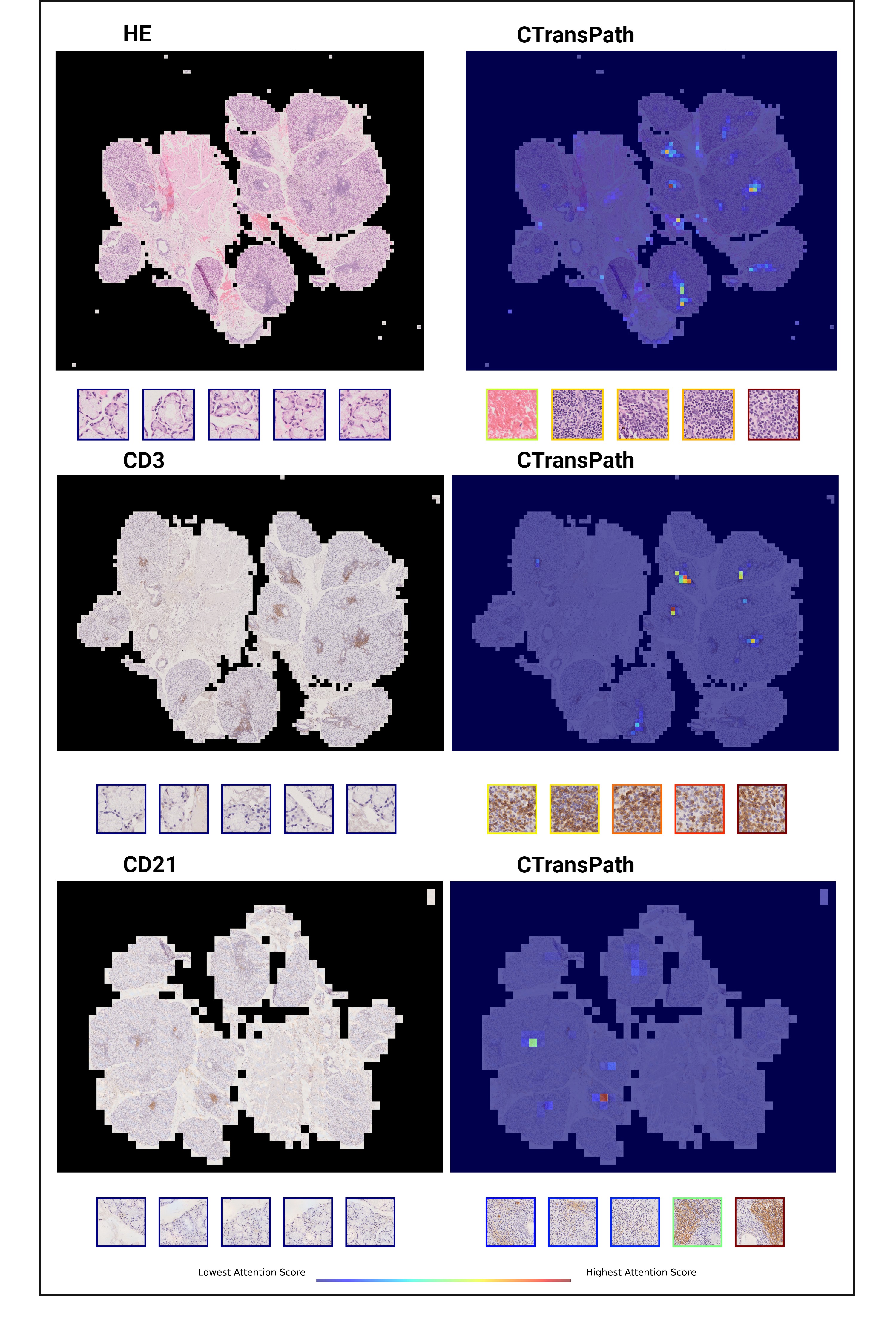}
    \caption{Attention heatmaps obtained from the ABMIL models trained with features extracted from CTransPath for a multi-stain (H\&E, CD3 and CD21) set of SD+ WSIs. The attention score pattern is very focused, with emphasis on a few well delimitated immune cell aggregates.}
    \label{ctranspath}
\end{figure}

\begin{figure}
    \centering
    \includegraphics[width=1\linewidth]{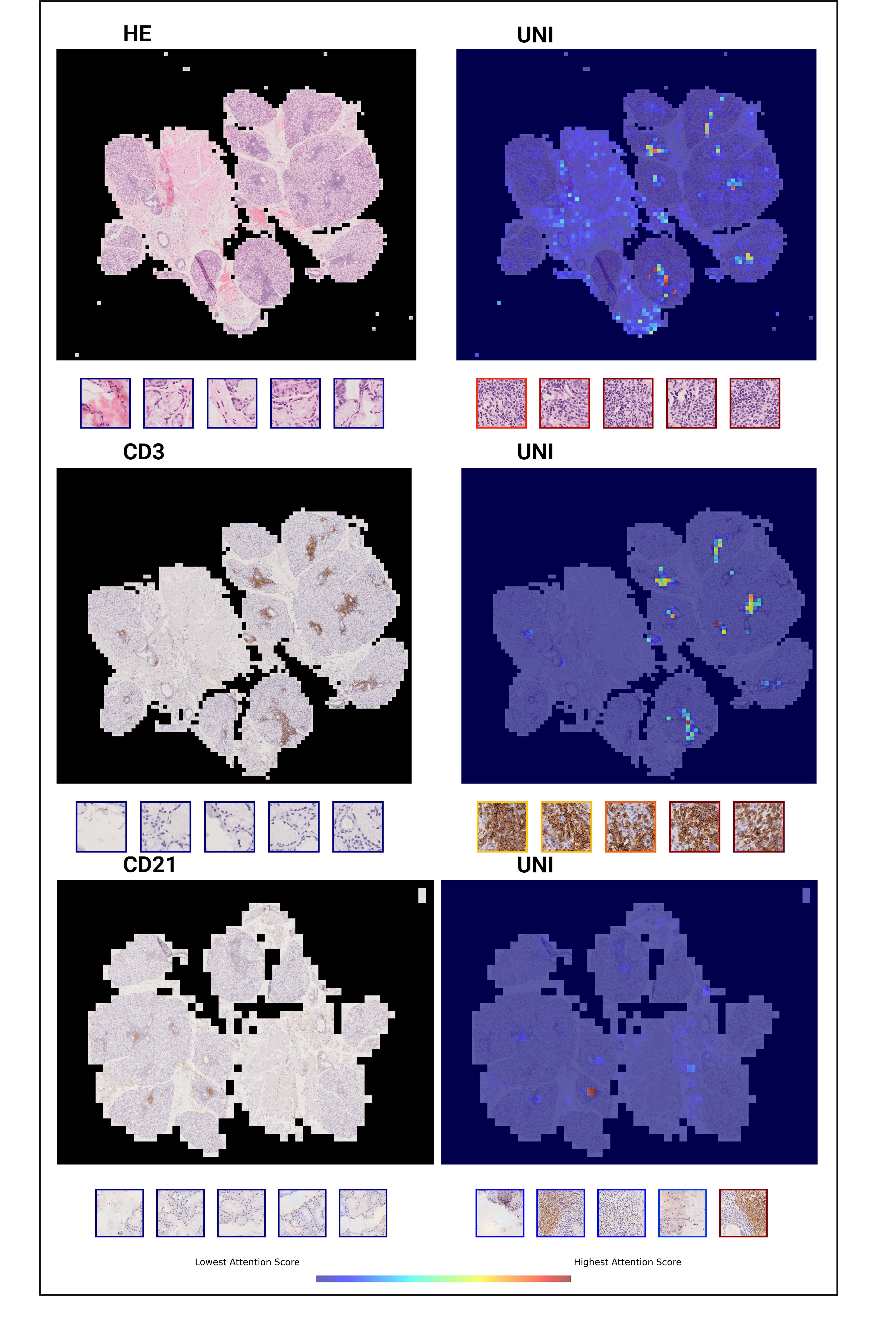}
    \caption{Attention heatmaps obtained from the ABMIL models trained with features extracted from UNI for a multi-stain (H\&E, CD3 and CD21) set of SD+ WSIs. Similarly to CTransPath, the attention score pattern is very focused, with emphasis on a few well delimitated immune cell aggregates.}
    \label{uni}
\end{figure}

\begin{figure}
    \centering
    \includegraphics[width=1\linewidth]{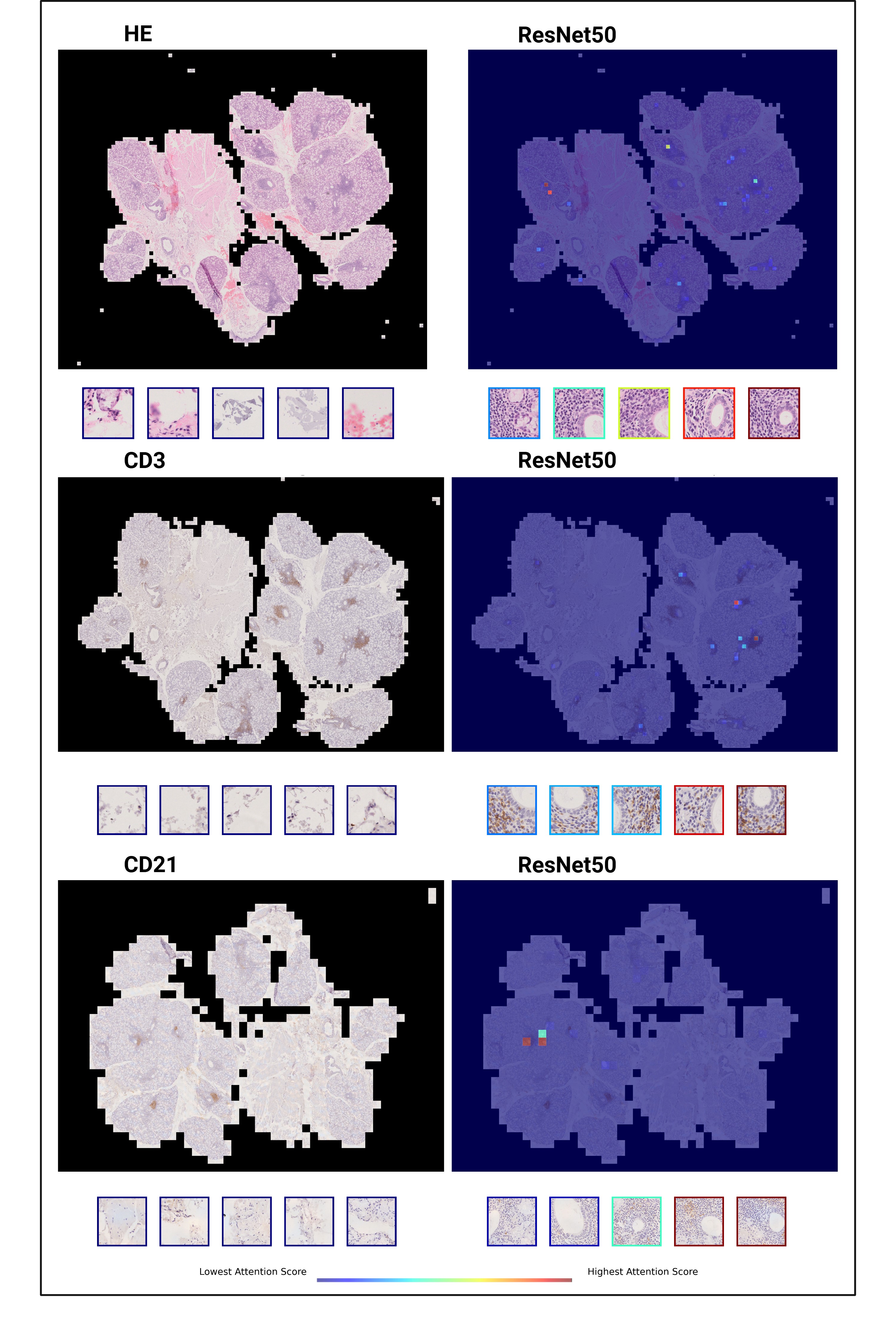}
    \caption{Attention heatmaps obtained from the ABMIL models trained with features extracted from ResNet50 for a multi-stain (H\&E, CD3 and CD21) set of SD+ WSIs. Attention scores patterns align with those seen in CTransPath \& UNI.}
    \label{resnet50}
\end{figure}

\begin{figure}
    \centering
    \includegraphics[width=1\linewidth]{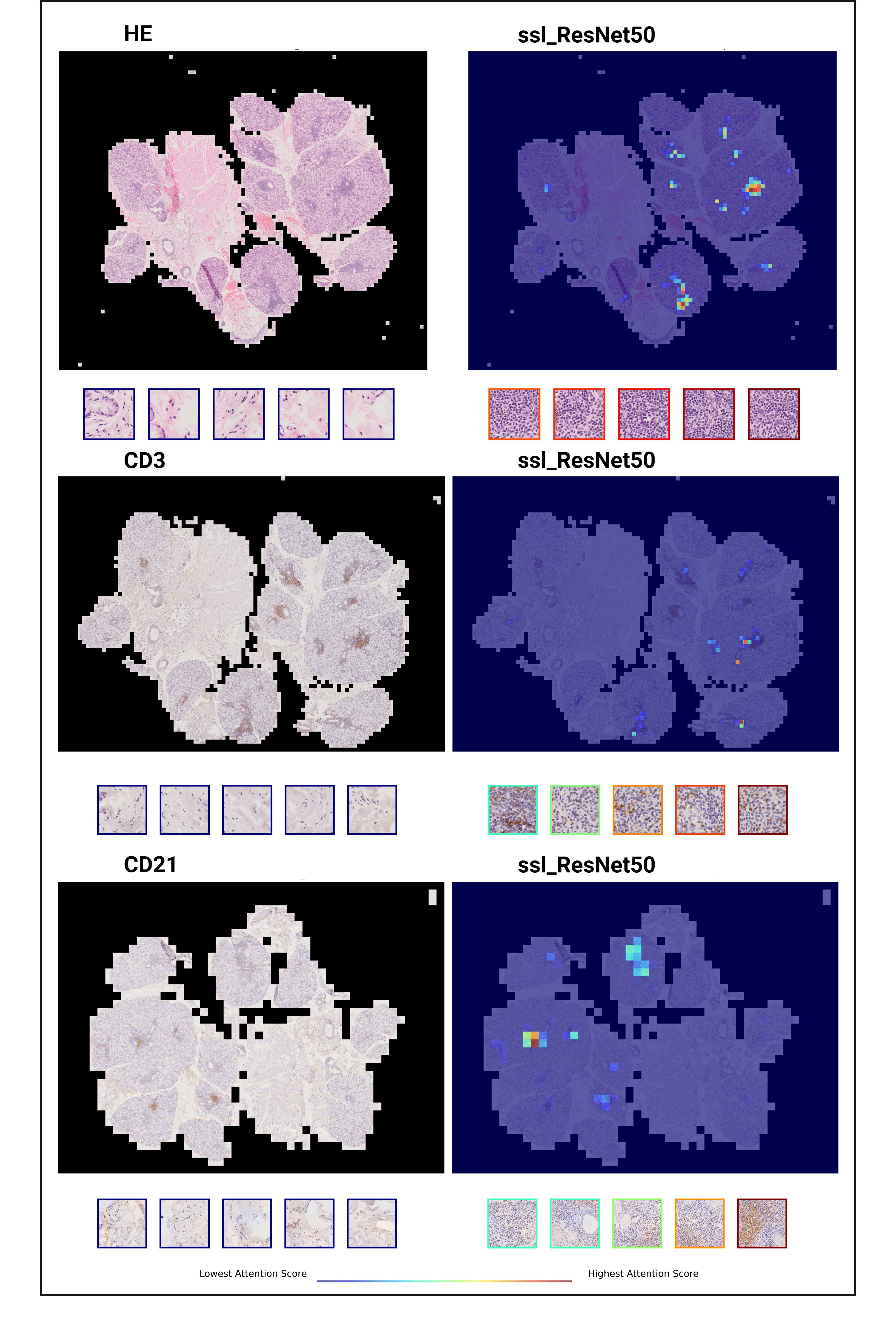}
    \caption{Attention heatmaps obtained from the ABMIL models trained with features extracted from ssl\_ResNet50 for a multi-stain (H\&E, CD3 and CD21) set of SD+ WSIs. Attention scores patterns align with those seen in CTransPath \& UNI.}
    \label{ssl_resnet50}
\end{figure}


\end{document}